\PassOptionsToPackage{table}{xcolor}

\documentclass{article} % For LaTeX2e
\usepackage{xcolor}
\usepackage{iclr2024_conference,times}

\usepackage{amsmath,amsfonts,bm}

\def\eqref#1{equation~\ref{#1}}
\def\1{\bm{1}}

\DeclareMathAlphabet{\mathsfit}{\encodingdefault}{\sfdefault}{m}{sl}
\SetMathAlphabet{\mathsfit}{bold}{\encodingdefault}{\sfdefault}{bx}{n}

\usepackage{hyperref}
\usepackage{url}

\usepackage{multirow}
\usepackage{booktabs}
\usepackage[T1]{fontenc}
\usepackage{subcaption}
\usepackage{ifthen}
\usepackage{pgfplots}
\pgfplotsset{compat=1.17}
\usepackage{amsmath}
\usepackage{multicol}
\usepackage[font=small,labelfont=bf]{caption}
\usepackage{pifont}

\def\rot{\rotatebox}

\definecolor{greenolive}{RGB}{46,154,85}
\definecolor{coolblue}{RGB}{64,193,200}

\title{Med42-v2: A Suite of Clinical LLMs}

\author{Clément Christophe\thanks{Equal contribution},~ Praveen K Kanithi\footnotemark[1],~ Tathagata Raha \\
\textbf{Shadab Khan,~ Marco AF Pimentel} \\
M42\\
Abu Dhabi,~ UAE\\
\texttt{\{cchristophe, pkanithi, traha, skhan, mpimentel\}@m42.ae} \\
}

\iclrfinalcopy % Uncomment for camera-ready version, but NOT for submission.
\begin{document}

\maketitle

\begin{abstract}
Med42-v2 introduces a suite of clinical large language models (LLMs) designed to address the limitations of generic models in healthcare settings. These models are built on Llama3 architecture and fine-tuned using specialized clinical data. They underwent multi-stage preference alignment to effectively respond to natural prompts. While generic models are often preference-aligned to avoid answering clinical queries as a precaution, Med42-v2 is specifically trained to overcome this limitation, enabling its use in clinical settings. Med42-v2 models demonstrate superior performance compared to the original Llama3 models in both 8B and 70B parameter configurations and GPT-4 across various medical benchmarks. These LLMs are developed to understand clinical queries, perform reasoning tasks, and provide valuable assistance in clinical environments. The models are now publicly available at \href{https://huggingface.co/m42-health}{https://huggingface.co/m42-health}.
\end{abstract}

\section{Introduction}
Large language models (LLMs) have revolutionized natural language processing, demonstrating remarkable capabilities across various domains \citep{achiam2023gpt, team2023gemini, anthropicIntroducingNext}. However, their application in specialized fields like healthcare has been limited due to the need for domain-specific knowledge and adherence to strict ethical and safety guidelines. The medical sector, in particular, requires models that can understand complex clinical terminology, reason through medical scenarios, and provide accurate, context-appropriate responses.

Despite the advancements, generic models face significant limitations when applied to healthcare settings. These include concerns about hallucinations and fabrications, biases and knowledge gaps, and risks about data privacy and ethics \citep{thirunavukarasu2023large, li2023ethics}. Such limitations reduce their effectiveness in aiding diagnostic processes \citep{de2023chatgpt, hirosawa2023diagnostic}, interpreting medical literature \citep{bagde2023systematic, cascella2023evaluating}, generating patient education materials \citep{ali2023using}, and assisting in clinical guidelines and decision support systems.
% (Souza et al. 2023; Hirosawa et al. 2023; Cascella et al. 2023; Bagde et al. 2023; Ali et al. 2023; Miner, Laranjo, and Kocaballi 2020).

To address these challenges, we introduce a second revision of Med42 \citep{christophe2024med42} called Med42-v2, a suite of clinical large language models designed to overcome the limitations of generic models in healthcare settings. Built on the Llama3 architecture \citep{dubey2024llama} and fine-tuned with specialized clinical data, Med42-v2 models undergo multi-stage preference alignment to effectively respond to natural prompts. Unlike generic models, which are often preference-aligned to avoid answering clinical queries as a precaution, Med42-v2 is specifically trained to engage with clinical queries, making it suitable for various stakeholders in healthcare, including clinicians, patients, and providers. Med42-v2 demonstrates superior performance compared to the original Llama3 models in both 8B and 70B parameter configurations (\autoref{tab:models}) across various medical benchmarks, excelling in understanding clinical queries, performing reasoning tasks, and providing valuable assistance in clinical environments.

The key contributions of this work are:
\begin{itemize}
    \item A suite of clinical LLMs (Med42-v2) built on Llama3 architecture, fine-tuned with specialized medical instruction data;
    \item A multi-stage preference alignment process to enhance the clinically fine-tuned model's ability to meet user expectations in healthcare settings;
    \item Empirical evidence demonstrating Med42-v2's superior performance over original Llama3 models in both 8B and 70B parameter configurations across various medical benchmarks.
\end{itemize}

\begin{table}[t!]
    \addtolength{\tabcolsep}{0pt}
    \def\arraystretch{1.2}
    \centering
    \small
    \begin{tabular}{llccc}
    \toprule
         & \textbf{Base Model} & \textbf{Finetuned} & \textbf{Aligned} & \textbf{Release Date} \\
    \midrule
         Med42-Llama2-70B & Llama2-70B & \textcolor{greenolive}{\ding{52}} & \textcolor{red}{\ding{56}} & October 2023 \\ 
         \rowcolor{coolblue!10} Med42-Llama3-8B\footnote{https://huggingface.co/m42-health/Llama3-Med42-8B} & Llama3-8B & \textcolor{greenolive}{\ding{52}} & \textcolor{greenolive}{\ding{52}} & June 2024 \\
         \rowcolor{coolblue!10} Med42-Llama3-70B\footnote{https://huggingface.co/m42-health/Llama3-Med42-70B} & Llama3-70B & \textcolor{greenolive}{\ding{52}} & \textcolor{greenolive}{\ding{52}} & June 2024 \\
         \rowcolor{coolblue!30} Med42-Llama3.1-8B & Llama3.1-8B & \textcolor{greenolive}{\ding{52}} & \textcolor{greenolive}{\ding{52}} & August 2024 \\
         \rowcolor{coolblue!30} Med42-Llama3.1-70B & Llama3.1-70B & \textcolor{greenolive}{\ding{52}} & \textcolor{greenolive}{\ding{52}} & August 2024 \\
    \bottomrule
    \end{tabular}
    \caption{Overview of Med42 and Med42-v2 suite of models.}
    \label{tab:models}
\end{table}

\section{Method}

The development of Med42-v2 follows a two-stage training process designed to create specialized clinical large language models (LLMs) that can effectively handle medical queries and tasks. Our approach builds upon the foundational capabilities of the Llama3 and Llama3.1 model families, enhancing them with domain-specific knowledge and alignment to clinical use cases.
Our training methodology consists of two primary stages:
\begin{itemize}
    \item \textbf{Instruction Fine-tuning:} In this initial stage, we fine-tune models from the Llama3 and Llama3.1 families using carefully curated clinical datasets. This process aims to improve the models with specialized medical knowledge.
    \item \textbf{Preference Alignment:} The second stage focuses on aligning the models' outputs with human preferences to ensure they can follow user instructions while safeguarding against unethical or biased behavior.
\end{itemize}

The following subsections detail each stage of our training process, highlighting the techniques and considerations involved in creating the Med42-v2 suite of models.

\subsection{Clinical Fine-Tuning Stage}
The clinical fine-tuning stage is an important step in adapting large language models for specialized medical applications. This phase aims to enhance the model's understanding and generation capabilities in clinical contexts, reducing the apprehension to answer medical-related questions and improving its relevance and accuracy for healthcare-related tasks. 

\paragraph{Datasets:}
To construct a training dataset tailored for clinical applications, we curated a diverse collection of resources specifically focused on medical and biomedical domains. Recognizing the importance of real-world usability beyond simple question-answering, we added examples demonstrating chain-of-thought reasoning as well as chat interactions. This addition was aimed at maximizing the model's reasoning capabilities and its effectiveness in conversational settings. To further enhance the model's generalizability and linguistic understanding, we incorporated a carefully selected subset of data from a general domain, comprising 26.5\% of the final training dataset. This hybrid approach is designed to optimize the model's performance across both specialized medical content and broader linguistic tasks.

\autoref{tab:traindata} provides a detailed breakdown of the various data subsets included in our study, along with their respective sample sizes.

\paragraph{Training Methodology:}
We employ the classic auto-regressive loss for fine-tuning. Loss is backpropagated only on output tokens. This approach ensures that the model learns to generate appropriate responses and not learn to generate the prompts. To maximize training speed and usage of the models context length, we concatenated all of our training samples into chunks of 8192 tokens.

\paragraph{Prompt Format:}
As we are fine-tuning the Instruct versions of Llama3 and Llama3.1, we adhere to their established prompt format, which includes system, assistant, and user fields.

\paragraph{Training Process:}
Each model was fine-tuned for two epochs over our curated dataset. The exact hyperparameters used in this process are detailed in \autoref{tab:hyperparameters}, providing full transparency for reproducibility.

\begin{table*}[t]
\centering
\small
\begin{tabular}{lcc}
\toprule
\textbf{Hyperparameter} & \textbf{Llama3/3.1 8B} & \textbf{Llama3/3.1 70B} \\
\midrule
GPU setup & 2 x 8 H100s & 6 x 8 H100s \\
LR scheduler & Linear warmup - Cosine & Linear warmup - Cosine\\
% & \multicolumn{3}{c|}{} & & & \\ % Add an empty row to adjust spacing
Maximum LR & $5\times10^{-6}$ & $5\times10^{-6}$ \\
Optimizer & AdamW & AdamW \\
Beta & (0.9, 0.95) & (0.9, 0.95) \\
% & \multicolumn{3}{c|}{} & & & \\ % Add an empty row to adjust spacing
Weight decay & 0.01 & 0.01 \\
% & \multicolumn{3}{c|}{} & & & \\ % Add an empty row to adjust spacing
Number of steps & 5,321 & 3,549  \\
Tokens per step & 262,144 & 393,216 \\
\bottomrule
\end{tabular}
\caption{Hyperparameters for the clinical fine-tuning stage}
\label{tab:hyperparameters}
\end{table*}

\subsection{Preference-Alignment Stage}
Preference alignment is a crucial step in developing large language models that can effectively meet user needs and expectations. This process involves adjusting the model's outputs to align with human preferences. However, obtaining direct human feedback at scale is challenging and resource-intensive. To address this, we employed open-access preference datasets created with AI feedback, allowing for more efficient and scalable alignment.

\paragraph{Datasets:} For our preference alignment phase, we utilized two primary datasets: the UltraFeedback dataset \citep{tunstall2023zephyr} and the Snorkel-DPO dataset \citep{huggingfaceSnorkelaiSnorkelMistralPairRMDPODatasetDatasets}. The UltraFeedback dataset is a comprehensive collection of AI preferences on various topics and tasks. The Snorkel-DPO dataset was created through an iterative process. Prompts were exclusively selected from UltraFeedback, without including external LLM responses. For each prompt, five response variations were generated using the Mistral-7B-Instruct-v0.2\footnote{\href{https://huggingface.co/mistralai/Mistral-7B-Instruct-v0.2}{https://huggingface.co/mistralai/Mistral-7B-Instruct-v0.2}} model. These responses were then reranked using PairRM \citep{llm-blender-2023} to identify the top (chosen) and bottom (rejected) responses. This process was repeated across three sets of 20,000 prompts, refining both the LLM and the dataset responses through three iterations. This method ensured a comprehensive and structured dataset for training purposes, improving with each iteration.

\paragraph{Training Methodology:} We employed Direct Preference Optimization (DPO) \citep{rafailov2024direct} to align our clinically fine-tuned checkpoints with preference data. This approach was chosen over more complex reinforcement learning algorithms \citep{ouyang2022training} due to its stability and scalability. We used DPO implementation from Huggingface Alignment Handbook library \citep{Tunstall_The_Alignment_Handbook} to train all our models.

\paragraph{Training Process:} We followed an iterative alignment approach \citep{viethoangtranduong} using the multi-stage data as described earlier. For the first iteration, we used UltraFeedback data and Snorkel-DPO-stage-1 data. The second and third iterations utilized Snorkel-DPO-stage-2 and Snorkel-DPO-stage-3 data, respectively. In each iteration, the model resulting from the previous iteration served as a reward model, leading to progressive performance improvements. The exact hyperparameters used in this process are detailed in \autoref{tab:hyperparameters_alignment}.

\begin{table*}[t]
\centering
\small
\begin{tabular}{lcc}
\toprule
\textbf{Hyperparameter} & \textbf{Llama3/3.1 8B} & \textbf{Llama3/3.1 70B} \\
\midrule
GPU setup & 2 x 8 H100s & 4 x 8 H100s \\
DPO-Beta & $0.1$ & $0.01$ \\
LR scheduler & Linear warmup - Cosine & Linear warmup - Cosine\\
% & \multicolumn{3}{c|}{} & & & \\ % Add an empty row to adjust spacing
Maximum LR & $1\times10^{-6}$ & $1\times10^{-6}$ \\
Optimizer & RMSprop & RMSprop \\
% & \multicolumn{3}{c|}{} & & & \\ % Add an empty row to adjust spacing
Weight decay & 0.0 & 0.0 \\
% & \multicolumn{3}{c|}{} & & & \\ % Add an empty row to adjust spacing
Batch size & 256 & 128  \\
Maximum length & 4096 & 4096 \\
Epochs (stage 1-3) & 1 & 1 \\
\bottomrule
\end{tabular}
\caption{Hyperparameters for the preference alignment stage. The same set of hyperparameters is consistently applied across all three stages of alignment.}
\label{tab:hyperparameters_alignment}
\end{table*}

\section{Benchmarks}
To assess the performance of the fine-tuned language models, following previous works \citep{singhal2023towards, chen2023meditron, toma2023clinical}, we used Eleuther AI’s evaluation harness framework \citep{eval-harness} to compute their zero-shot performance across various commonly-used medical benchmarks. These contain medical exam questions and research datasets with multiple-choice answers, and include: MMLU (medical subset) \citep{hendryckstest2021}, MMLU-Pro \citep{Wang2024ds}, MedMCQA \citep{pmlr-v174-pal22a}, MedQA \citep{jin2020disease}, USMLE \citep{nori2023capabilities, Han2023medalpaca}, PubmedQA \citep{jin2019pubmedqa}, ToxiGen \citep{hartvigsen2022toxigen}. All datasets are in the English language and all questions containing images were excluded. The harness framework has been updated to include chat templates. Additionally, our log-likelihood calculations are over the entire response sequence instead of just the first token.

\autoref{tab:closed-ended-results} presents a comparison of our medically-aligned models with various clinical and general-purpose LLMs. The latest version of Med42 shows significant improvements over its previous iteration across multiple benchmarks. In particular, the larger Med42 models (70B) exceed the performance of other leading general-purpose and domain-specific models, even outperforming proprietary models like GPT-4.0 \citep{achiam2023gpt} on all datasets. This suggests that targeted medical instruction and alignment enhance the model's clinical knowledge and reasoning abilities.

These findings consistently demonstrate that larger models perform better on these tasks, in line with general trends in language model scaling. However, the performance gains are less significant on safety-focused benchmarks like ToxiGen. Moreover, models such as Med42 and OpenBioLLM exhibit enhanced performance on these benchmarks compared to their base Llama3-Instruct versions. This highlights the advantages of specific medical instruction and alignment in improving the models' clinical expertise and analytical capabilities.

It's worth noting that these results represent zero-shot performance. Prior research has indicated that prompting techniques, such as \emph{Medprompt} \citep{Nori2023vc}, or integration with search functionalities can yield even higher accuracy rates. For instance, Med-Gemini has achieved a 91.2\% accuracy on benchmarks like MedQA \citep{Saab2024yx}.

LLMs are designed to excel across a diverse set of tasks, leveraging their conversational capabilities. This versatility is crucial for their application in various clinical tasks. Our future work will focus on evaluating these capabilities in a clinical setting in detail.

\begin{table}[t!]
\addtolength{\tabcolsep}{0pt}
\def\arraystretch{1.1}
\centering
\small
\begin{tabular}{lccccccc|c}
\toprule
\textbf{Model} & \rot{60}{\textbf{MMLU-Pro}} & \rot{60}{\textbf{MMLU}} & \rot{65}{\textbf{MedMCQA}} & \rot{60}{\textbf{MedQA}} & \rot{60}{\textbf{USMLE}} & \rot{60}{\textbf{PubmedQA}} & \rot{60}{\textbf{ToxiGen}} & \textbf{Avg.}\\
\midrule
Mistral-7B-Instruct-v0.3 & 33.8 & 64.6 & 46.3 & 49.3 & 50.4 & 42.8 & \textbf{86.2} & 53.3 \\
Llama3-8B-Instruct  & 48.2 & 72.9 & 59.7 & 61.6 & 60.4 & 69.8 & 78.5 & 64.4 \\
Llama3.1-8B-Instruct & 49.9  & 73.4 & 58.4 & 62.0 & \underline{68.2} & \textbf{76.2} & 82.3 & 67.2 \\
JSL-MedLlama-3-8B-v2.0 & 46.9 & \textbf{75.9} & 59.7 & 59.9 & 60.6 & \underline{75.0} & 74.3 & 64.6 \\
\rowcolor{coolblue!10} Med42-Llama3-8B & \textbf{54.3} & \underline{75.8} & \textbf{61.3} & \underline{62.8} & 67.0 & 68.4 & 81.5 & \underline{67.3} \\
\rowcolor{coolblue!30} Med42-Llama3.1-8B & \underline{54.2} & 73.6 & \underline{59.7} & \textbf{63.2} & \textbf{69.9} & 72.2 & \underline{83.8} & \textbf{68.1} \\						
\midrule
Gemma-2-9B & 49.9 & 78.8 & 56.2 & 60.9 & 66.8 & 39.4 & 70.5 & 60.4 \\
Falcon-11B &  26.3  & 62.2 &  43.8  & 43.1 & 44.1 & 58.0 & 68.9 & 49.5 \\
Gemma-2-27B & \underline{55.8} & \underline{81.3} & 60.2 & \underline{65.7} & \underline{71.5} & 51.4 & 69.3 & \underline{65.0} \\
Mixtral-8x7B-Instruct & 46.9 & 75.6 & 54.1 & 58.4 & 67.1 & \underline{63.2} & \underline{83.5} & 64.1 \\
BiMediX (Eng) & 49.7 & 74.9 & \underline{61.1} & 65.1 & 66.4 & \textbf{77.8} & 43.2 & 62.6 \\
Phi-3-Medium-128k-instruct & \textbf{58.2} & \textbf{81.4} & \textbf{61.5} & \textbf{69.0} & \textbf{73.9} & 46.4 & \textbf{86.6} & \textbf{68.1} \\
\midrule
Mixtral-8x22B-Instruct & 55.6 & 80.7 & 61.4 & 67.2 & 76.1 & 62.2 & 87.1 & 70.0 \\
Llama3-70B-Instruct & 64.2 & 86.0 & 72.0 & 78.9 & 83.6 & 71.8 & 87.6 & 77.7 \\
Llama3.1-70B-Instruct & \underline{64.6} & \underline{87.4} & 71.9 & 78.6 & \underline{93.4} & 76.6 & \textbf{91.3} & \underline{80.5} \\
OpenBioLLM-70B & 64.2 & \textbf{90.4} & \textbf{73.2} & 76.9 & 79.0 & 73.2 & \textbf{91.3} & 78.3 \\
Med42-Llama2-70B & 51.5 & 76.7 & 60.9 & 61.5 & 71.9 & 64.6 & 88.8 & 68.0 \\
\rowcolor{coolblue!10} Med42-Llama3-70B & 64.4 & 87.1 & \textbf{73.2} & \underline{79.1} & 83.8 & \textbf{78.8} & 90.3 & 79.5 \\
\rowcolor{coolblue!30} Med42-Llama3.1-70B &  \textbf{66.1} & 86.8 & \underline{72.4} & \textbf{80.4} & \textbf{94.5} & \underline{77.6} & \underline{90.4} & \textbf{81.2} \\
\midrule
Mistral-Large-Instruct-2407 & \underline{66.4} & \underline{87.5} & 68.3 & 75.9 & \underline{85.8} & 56.2 & \textbf{91.1} & 75.9 \\
GPT-4.0\textsuperscript{\rm \textdagger}     & -    & 87.0 & \underline{69.5} & \underline{78.9} & 84.1 & \textbf{75.2} & - & \underline{78.9} \\
Llama3.1-405B-Instruct & \textbf{70.2} & \textbf{89.3} & \textbf{75.8} & \textbf{81.9} & \textbf{95.5} & \underline{74.6} & \underline{90.7} & \textbf{82.6} \\
% Med-Gemini   & -    & -    &  - & \textbf{84.0} & - & & - \\
% Med-PaLM-2 (5-shot)* & - & \underline{87.8} & 71.3 & 79.7 & - & & - \\
\bottomrule
\end{tabular}
\caption{Performance of Med42-v2 models on key closed-ended medical benchmark (zero-shot) evaluations. We compare the performance with that of competing models, and we \textbf{boldface} and \underline{underline} the best and second best-performing models (respectively) in each of the three model-size equivalence classes. \textsuperscript{\textdagger}Performance results for GPT-4.0 have been reported in \cite{nori2023capabilities}. %Med42-v2, built upon Llama3, shows significant improvements across all benchmarks, demonstrating state-of-the-art results in closed-ended medical tasks. Notably, Med42v2 surpasses even closed-source models like GPT-4.0 on several benchmarks.
}
\label{tab:closed-ended-results}
\end{table}

\section{Conclusions and Limitations}
In conclusion, we introduced Med42-v2, a suite of clinical large language models built on the Llama3 architecture and fine-tuned with specialized clinical data. Med42-v2 also employs a multi-stage preference alignment process, enabling it to effectively handle clinical queries. Our empirical results show that Med42-v2 outperforms the original Llama3 models in both 8B and 70B parameter configurations and GPT-4 across various medical benchmarks.

However, utilizing clinical LLMs in real-world settings can present several limitations. Despite improvements, Med42-v2 may not entirely be free from issues like hallucinations, biases, and ethical concerns, which are particularly critical in the medical field. The reliance on high-quality, domain-specific data means that any gaps or biases in the training data could impact the model's effectiveness. To address these concerns, our future work involves developing a new evaluation framework to assess the clinical utility of LLMs by testing them on real-world use cases. This framework will focus on evaluating clinical data understanding, safety, and reasoning capabilities, providing a more comprehensive understanding of how these models perform in practical, high-stakes environments. By rigorously testing LLMs in real-world scenarios, we aim to identify and mitigate potential risks, ensuring that models like Med42-v2 can be safely and effectively integrated into healthcare settings.



\bibliography{iclr2024_conference}

\begin{thebibliography}{49}
\providecommand{\natexlab}[1]{#1}
\providecommand{\url}[1]{\texttt{#1}}
\expandafter\ifx\csname urlstyle\endcsname\relax
  \providecommand{\doi}[1]{doi: #1}\else
  \providecommand{\doi}{doi: \begingroup \urlstyle{rm}\Url}\fi

\bibitem[Achiam et~al.(2023)Achiam, Adler, Agarwal, Ahmad, Akkaya, Aleman, Almeida, Altenschmidt, Altman, Anadkat, et~al.]{achiam2023gpt}
Josh Achiam, Steven Adler, Sandhini Agarwal, Lama Ahmad, Ilge Akkaya, Florencia~Leoni Aleman, Diogo Almeida, Janko Altenschmidt, Sam Altman, Shyamal Anadkat, et~al.
\newblock Gpt-4 technical report.
\newblock \emph{arXiv preprint arXiv:2303.08774}, 2023.

\bibitem[Ali et~al.(2023)Ali, Dobbs, Hutchings, and Whitaker]{ali2023using}
Stephen~R Ali, Thomas~D Dobbs, Hayley~A Hutchings, and Iain~S Whitaker.
\newblock Using chatgpt to write patient clinic letters.
\newblock \emph{The Lancet Digital Health}, 5\penalty0 (4):\penalty0 e179--e181, 2023.

\bibitem[Altaf(2023)]{medical-instruction-120k}
Mohammed Altaf.
\newblock Medical instruction 120k, 2023.
\newblock URL \url{https://t.ly/Pn5lg}.

\bibitem[Anthropic(2024)]{anthropicIntroducingNext}
Anthropic.
\newblock {I}ntroducing the next generation of {C}laude --- anthropic.com.
\newblock \url{https://t.ly/JQLxz}, 2024.
\newblock [Accessed 09-07-2024].

\bibitem[Bagde et~al.(2023)Bagde, Dhopte, Alam, and Basri]{bagde2023systematic}
Hiroj Bagde, Ashwini Dhopte, Mohammad~Khursheed Alam, and Rehana Basri.
\newblock A systematic review and meta-analysis on chatgpt and its utilization in medical and dental research.
\newblock \emph{Heliyon}, 9\penalty0 (12), 2023.

\bibitem[{Ben Abacha} \& Demner{-}Fushman(2019){Ben Abacha} and Demner{-}Fushman]{BenAbacha-BMC-2019}
Asma {Ben Abacha} and Dina Demner{-}Fushman.
\newblock A question-entailment approach to question answering.
\newblock \emph{{BMC} Bioinform.}, 20\penalty0 (1):\penalty0 511:1--511:23, 2019.
\newblock URL \url{https://t.ly/U1FFw}.

\bibitem[Ben~Abacha et~al.(2023)Ben~Abacha, Yim, Fan, and Lin]{mts-dialog}
Asma Ben~Abacha, Wen-wai Yim, Yadan Fan, and Thomas Lin.
\newblock An empirical study of clinical note generation from doctor-patient encounters.
\newblock In \emph{Proceedings of the 17th Conference of the European Chapter of the Association for Computational Linguistics}, pp.\  2291--2302, Dubrovnik, Croatia, May 2023. Association for Computational Linguistics.
\newblock URL \url{https://aclanthology.org/2023.eacl-main.168}.

\bibitem[Cascella et~al.(2023)Cascella, Montomoli, Bellini, and Bignami]{cascella2023evaluating}
Marco Cascella, Jonathan Montomoli, Valentina Bellini, and Elena Bignami.
\newblock Evaluating the feasibility of chatgpt in healthcare: an analysis of multiple clinical and research scenarios.
\newblock \emph{Journal of medical systems}, 47\penalty0 (1):\penalty0 33, 2023.

\bibitem[Chen et~al.(2023)Chen, Cano, Romanou, Bonnet, Matoba, Salvi, Pagliardini, Fan, K{\"o}pf, Mohtashami, et~al.]{chen2023meditron}
Zeming Chen, Alejandro~Hern{\'a}ndez Cano, Angelika Romanou, Antoine Bonnet, Kyle Matoba, Francesco Salvi, Matteo Pagliardini, Simin Fan, Andreas K{\"o}pf, Amirkeivan Mohtashami, et~al.
\newblock Meditron-70b: Scaling medical pretraining for large language models.
\newblock \emph{arXiv preprint arXiv:2311.16079}, 2023.

\bibitem[Christophe et~al.(2024)Christophe, Kanithi, Munjal, Raha, Hayat, Rajan, Al-Mahrooqi, Gupta, Salman, Gosal, et~al.]{christophe2024med42}
Cl{\'e}ment Christophe, Praveen~K Kanithi, Prateek Munjal, Tathagata Raha, Nasir Hayat, Ronnie Rajan, Ahmed Al-Mahrooqi, Avani Gupta, Muhammad~Umar Salman, Gurpreet Gosal, et~al.
\newblock Med42--evaluating fine-tuning strategies for medical llms: Full-parameter vs. parameter-efficient approaches.
\newblock \emph{arXiv preprint arXiv:2404.14779}, 2024.

\bibitem[de~Souza et~al.(2023)de~Souza, Fonseca, Martins, de~Almeida, Pontes, Coracin, Lopes, Khurram, Santos-Silva, Hagag, et~al.]{de2023chatgpt}
Lucas~Lacerda de~Souza, Felipe~Paiva Fonseca, Manoela~Domingues Martins, Oslei~Paes de~Almeida, Helder Ant{\^o}nio~Rebelo Pontes, F{\'a}bio~Luiz Coracin, M{\'a}rcio~Ajudarte Lopes, Syed~Ali Khurram, Alan~Roger Santos-Silva, Ahmed Hagag, et~al.
\newblock Chatgpt and medicine: a potential threat to science or a step towards the future?
\newblock \emph{Journal of Medical Artificial Intelligence}, 6, 2023.

\bibitem[Ding et~al.(2023)Ding, Chen, Xu, Qin, Zheng, Hu, Liu, Sun, and Zhou]{ding2023enhancing}
Ning Ding, Yulin Chen, Bokai Xu, Yujia Qin, Zhi Zheng, Shengding Hu, Zhiyuan Liu, Maosong Sun, and Bowen Zhou.
\newblock Enhancing chat language models by scaling high-quality instructional conversations, 2023.

\bibitem[Dubey et~al.(2024)Dubey, Jauhri, Pandey, Kadian, Al-Dahle, Letman, Mathur, Schelten, Yang, Fan, et~al.]{dubey2024llama}
Abhimanyu Dubey, Abhinav Jauhri, Abhinav Pandey, Abhishek Kadian, Ahmad Al-Dahle, Aiesha Letman, Akhil Mathur, Alan Schelten, Amy Yang, Angela Fan, et~al.
\newblock The llama 3 herd of models.
\newblock \emph{arXiv preprint arXiv:2407.21783}, 2024.

\bibitem[Gao et~al.(2023)Gao, Tow, Abbasi, Biderman, Black, DiPofi, Foster, Golding, Hsu, Le~Noac'h, Li, McDonell, Muennighoff, Ociepa, Phang, Reynolds, Schoelkopf, Skowron, Sutawika, Tang, Thite, Wang, Wang, and Zou]{eval-harness}
Leo Gao, Jonathan Tow, Baber Abbasi, Stella Biderman, Sid Black, Anthony DiPofi, Charles Foster, Laurence Golding, Jeffrey Hsu, Alain Le~Noac'h, Haonan Li, Kyle McDonell, Niklas Muennighoff, Chris Ociepa, Jason Phang, Laria Reynolds, Hailey Schoelkopf, Aviya Skowron, Lintang Sutawika, Eric Tang, Anish Thite, Ben Wang, Kevin Wang, and Andy Zou.
\newblock A framework for few-shot language model evaluation, 12 2023.
\newblock URL \url{https://zenodo.org/records/10256836}.

\bibitem[Gemini et~al.(2023)Gemini, Anil, Borgeaud, Wu, Alayrac, Yu, Soricut, Schalkwyk, Dai, Hauth, et~al.]{team2023gemini}
Team Gemini, Rohan Anil, Sebastian Borgeaud, Yonghui Wu, Jean-Baptiste Alayrac, Jiahui Yu, Radu Soricut, Johan Schalkwyk, Andrew~M Dai, Anja Hauth, et~al.
\newblock Gemini: a family of highly capable multimodal models.
\newblock \emph{arXiv preprint arXiv:2312.11805}, 2023.

\bibitem[Han et~al.(2023)Han, Adams, Papaioannou, Grundmann, Oberhauser, L{\"o}ser, Truhn, and Bressem]{Han2023medalpaca}
Tianyu Han, Lisa~C Adams, Jens-Michalis Papaioannou, Paul Grundmann, Tom Oberhauser, Alexander L{\"o}ser, Daniel Truhn, and Keno~K Bressem.
\newblock Medalpaca--an open-source collection of medical conversational ai models and training data.
\newblock \emph{arXiv preprint arXiv:2304.08247}, 2023.

\bibitem[Hartvigsen et~al.(2022)Hartvigsen, Gabriel, Palangi, Sap, Ray, and Kamar]{hartvigsen2022toxigen}
Thomas Hartvigsen, Saadia Gabriel, Hamid Palangi, Maarten Sap, Dipankar Ray, and Ece Kamar.
\newblock Toxigen: A large-scale machine-generated dataset for adversarial and implicit hate speech detection.
\newblock \emph{arXiv preprint arXiv:2203.09509}, 2022.

\bibitem[Hendrycks et~al.(2021)Hendrycks, Burns, Basart, Zou, Mazeika, Song, and Steinhardt]{hendryckstest2021}
Dan Hendrycks, Collin Burns, Steven Basart, Andy Zou, Mantas Mazeika, Dawn Song, and Jacob Steinhardt.
\newblock Measuring massive multitask language understanding.
\newblock \emph{Proceedings of the International Conference on Learning Representations (ICLR)}, 2021.

\bibitem[Hirosawa et~al.(2023)Hirosawa, Harada, Yokose, Sakamoto, Kawamura, and Shimizu]{hirosawa2023diagnostic}
Takanobu Hirosawa, Yukinori Harada, Masashi Yokose, Tetsu Sakamoto, Ren Kawamura, and Taro Shimizu.
\newblock Diagnostic accuracy of differential-diagnosis lists generated by generative pretrained transformer 3 chatbot for clinical vignettes with common chief complaints: a pilot study.
\newblock \emph{International journal of environmental research and public health}, 20\penalty0 (4):\penalty0 3378, 2023.

\bibitem[Jiang et~al.(2023)Jiang, Ren, and Lin]{llm-blender-2023}
Dongfu Jiang, Xiang Ren, and Bill~Yuchen Lin.
\newblock Llm-blender: Ensembling large language models with pairwise comparison and generative fusion.
\newblock In \emph{Proceedings of the 61th Annual Meeting of the Association for Computational Linguistics (ACL 2023)}, 2023.

\bibitem[Jin et~al.(2020)Jin, Pan, Oufattole, Weng, Fang, and Szolovits]{jin2020disease}
Di~Jin, Eileen Pan, Nassim Oufattole, Wei-Hung Weng, Hanyi Fang, and Peter Szolovits.
\newblock What disease does this patient have? a large-scale open domain question answering dataset from medical exams.
\newblock \emph{arXiv preprint arXiv:2009.13081}, 2020.

\bibitem[Jin et~al.(2019)Jin, Dhingra, Liu, Cohen, and Lu]{jin2019pubmedqa}
Qiao Jin, Bhuwan Dhingra, Zhengping Liu, William Cohen, and Xinghua Lu.
\newblock Pubmedqa: A dataset for biomedical research question answering.
\newblock In \emph{Proceedings of the 2019 Conference on Empirical Methods in Natural Language Processing and the 9th International Joint Conference on Natural Language Processing (EMNLP-IJCNLP)}, pp.\  2567--2577, 2019.

\bibitem[Johannes~Welbl(2017)]{SciQ}
Matt~Gardner Johannes~Welbl, Nelson F.~Liu.
\newblock Crowdsourcing multiple choice science questions.
\newblock 2017.

\bibitem[Kotonya \& Toni(2020)Kotonya and Toni]{kotonya2020explainable}
Neema Kotonya and Francesca Toni.
\newblock Explainable automated fact-checking for public health claims.
\newblock \emph{arXiv preprint arXiv:2010.09926}, 2020.

\bibitem[Lambert et~al.(2023)Lambert, Tunstall, Rajani, and Thrush]{h4stackexchange}
Nathan Lambert, Lewis Tunstall, Nazneen Rajani, and Tristan Thrush.
\newblock Huggingface h4 stack exchange preference dataset, 2023.
\newblock URL \url{https://t.ly/2sE9E}.

\bibitem[Li et~al.(2023)Li, Moon, Purkayastha, Celi, Trivedi, and Gichoya]{li2023ethics}
Hanzhou Li, John~T Moon, Saptarshi Purkayastha, Leo~Anthony Celi, Hari Trivedi, and Judy~W Gichoya.
\newblock Ethics of large language models in medicine and medical research.
\newblock \emph{The Lancet Digital Health}, 5\penalty0 (6):\penalty0 e333--e335, 2023.

\bibitem[Lian et~al.(2023)Lian, Goodson, Pentland, Cook, Vong, and "Teknium"]{OpenOrca}
Wing Lian, Bleys Goodson, Eugene Pentland, Austin Cook, Chanvichet Vong, and "Teknium".
\newblock Openorca: An open dataset of gpt augmented flan reasoning traces.
\newblock \url{https://https://huggingface.co/Open-Orca/OpenOrca}, 2023.

\bibitem[Longpre et~al.(2023)Longpre, Hou, Vu, Webson, Chung, Tay, Zhou, Le, Zoph, Wei, et~al.]{longpre2023flan}
Shayne Longpre, Le~Hou, Tu~Vu, Albert Webson, Hyung~Won Chung, Yi~Tay, Denny Zhou, Quoc~V Le, Barret Zoph, Jason Wei, et~al.
\newblock The flan collection: Designing data and methods for effective instruction tuning.
\newblock \emph{arXiv preprint arXiv:2301.13688}, 2023.

\bibitem[Nori et~al.(2023{\natexlab{a}})Nori, King, McKinney, Carignan, and Horvitz]{nori2023capabilities}
Harsha Nori, Nicholas King, Scott~Mayer McKinney, Dean Carignan, and Eric Horvitz.
\newblock Capabilities of gpt-4 on medical challenge problems, 2023{\natexlab{a}}.

\bibitem[Nori et~al.(2023{\natexlab{b}})Nori, Lee, Zhang, Carignan, Edgar, Fusi, King, Larson, Li, Liu, Luo, McKinney, Ness, Poon, Qin, Usuyama, White, and Horvitz]{Nori2023vc}
Harsha Nori, Yin~Tat Lee, Sheng Zhang, Dean Carignan, Richard Edgar, Nicolo Fusi, Nicholas King, Jonathan Larson, Yuanzhi Li, Weishung Liu, Renqian Luo, Scott~Mayer McKinney, Robert~Osazuwa Ness, Hoifung Poon, Tao Qin, Naoto Usuyama, Chris White, and Eric Horvitz.
\newblock Can generalist foundation models outcompete {Special-Purpose} tuning? case study in medicine.
\newblock November 2023{\natexlab{b}}.

\bibitem[OpenChat(2023)]{cogstack-opengpt-sharegpt}
OpenChat, 2023.
\newblock URL \url{https://t.ly/UN-Hu}.

\bibitem[Ouyang et~al.(2022)Ouyang, Wu, Jiang, Almeida, Wainwright, Mishkin, Zhang, Agarwal, Slama, Ray, et~al.]{ouyang2022training}
Long Ouyang, Jeffrey Wu, Xu~Jiang, Diogo Almeida, Carroll Wainwright, Pamela Mishkin, Chong Zhang, Sandhini Agarwal, Katarina Slama, Alex Ray, et~al.
\newblock Training language models to follow instructions with human feedback.
\newblock \emph{Advances in neural information processing systems}, 35:\penalty0 27730--27744, 2022.

\bibitem[Pal et~al.(2022)Pal, Umapathi, and Sankarasubbu]{pmlr-v174-pal22a}
Ankit Pal, Logesh~Kumar Umapathi, and Malaikannan Sankarasubbu.
\newblock Medmcqa: A large-scale multi-subject multi-choice dataset for medical domain question answering.
\newblock In Gerardo Flores, George~H Chen, Tom Pollard, Joyce~C Ho, and Tristan Naumann (eds.), \emph{Proceedings of the Conference on Health, Inference, and Learning}, volume 174 of \emph{Proceedings of Machine Learning Research}, pp.\  248--260. PMLR, 07--08 Apr 2022.
\newblock URL \url{https://proceedings.mlr.press/v174/pal22a.html}.

\bibitem[Rafailov et~al.(2024)Rafailov, Sharma, Mitchell, Manning, Ermon, and Finn]{rafailov2024direct}
Rafael Rafailov, Archit Sharma, Eric Mitchell, Christopher~D Manning, Stefano Ermon, and Chelsea Finn.
\newblock Direct preference optimization: Your language model is secretly a reward model.
\newblock \emph{Advances in Neural Information Processing Systems}, 36, 2024.

\bibitem[Saab et~al.(2024)Saab, Tu, Weng, Tanno, Stutz, Wulczyn, Zhang, Strother, Park, Vedadi, Chaves, Hu, Schaekermann, Kamath, Cheng, Barrett, Cheung, Mustafa, Palepu, McDuff, Hou, Golany, Liu, Alayrac, Houlsby, Tomasev, Freyberg, Lau, Kemp, Lai, Azizi, Kanada, Man, Kulkarni, Sun, Shakeri, He, Caine, Webson, Latysheva, Johnson, Mansfield, Lu, Rivlin, Anderson, Green, Wong, Krause, Shlens, Dominowska, Ali~Eslami, Chou, Cui, Vinyals, Kavukcuoglu, Manyika, Dean, Hassabis, Matias, Webster, Barral, Corrado, Semturs, Sara~Mahdavi, Gottweis, Karthikesalingam, and Natarajan]{Saab2024yx}
Khaled Saab, Tao Tu, Wei-Hung Weng, Ryutaro Tanno, David Stutz, Ellery Wulczyn, Fan Zhang, Tim Strother, Chunjong Park, Elahe Vedadi, Juanma~Zambrano Chaves, Szu-Yeu Hu, Mike Schaekermann, Aishwarya Kamath, Yong Cheng, David G~T Barrett, Cathy Cheung, Basil Mustafa, Anil Palepu, Daniel McDuff, Le~Hou, Tomer Golany, Luyang Liu, Jean-Baptiste Alayrac, Neil Houlsby, Nenad Tomasev, Jan Freyberg, Charles Lau, Jonas Kemp, Jeremy Lai, Shekoofeh Azizi, Kimberly Kanada, Siwai Man, Kavita Kulkarni, Ruoxi Sun, Siamak Shakeri, Luheng He, Ben Caine, Albert Webson, Natasha Latysheva, Melvin Johnson, Philip Mansfield, Jian Lu, Ehud Rivlin, Jesper Anderson, Bradley Green, Renee Wong, Jonathan Krause, Jonathon Shlens, Ewa Dominowska, S~M Ali~Eslami, Katherine Chou, Claire Cui, Oriol Vinyals, Koray Kavukcuoglu, James Manyika, Jeff Dean, Demis Hassabis, Yossi Matias, Dale Webster, Joelle Barral, Greg Corrado, Christopher Semturs, S~Sara~Mahdavi, Juraj Gottweis, Alan Karthikesalingam, and Vivek Natarajan.
\newblock Capabilities of gemini models in medicine.
\newblock April 2024.

\bibitem[Sanh et~al.(2022)Sanh, Webson, Raffel, Bach, Sutawika, Alyafeai, Chaffin, Stiegler, Scao, Raja, Dey, Bari, Xu, Thakker, Sharma, Szczechla, Kim, Chhablani, Nayak, Datta, Chang, Jiang, Wang, Manica, Shen, Yong, Pandey, Bawden, Wang, Neeraj, Rozen, Sharma, Santilli, Fevry, Fries, Teehan, Bers, Biderman, Gao, Wolf, and Rush]{sanh2022multitask}
Victor Sanh, Albert Webson, Colin Raffel, Stephen~H. Bach, Lintang Sutawika, Zaid Alyafeai, Antoine Chaffin, Arnaud Stiegler, Teven~Le Scao, Arun Raja, Manan Dey, M~Saiful Bari, Canwen Xu, Urmish Thakker, Shanya~Sharma Sharma, Eliza Szczechla, Taewoon Kim, Gunjan Chhablani, Nihal Nayak, Debajyoti Datta, Jonathan Chang, Mike Tian-Jian Jiang, Han Wang, Matteo Manica, Sheng Shen, Zheng~Xin Yong, Harshit Pandey, Rachel Bawden, Thomas Wang, Trishala Neeraj, Jos Rozen, Abheesht Sharma, Andrea Santilli, Thibault Fevry, Jason~Alan Fries, Ryan Teehan, Tali Bers, Stella Biderman, Leo Gao, Thomas Wolf, and Alexander~M. Rush.
\newblock Multitask prompted training enables zero-shot task generalization, 2022.

\bibitem[Singhal et~al.(2023)Singhal, Tu, Gottweis, Sayres, Wulczyn, Hou, Clark, Pfohl, Cole-Lewis, Neal, et~al.]{singhal2023towards}
Karan Singhal, Tao Tu, Juraj Gottweis, Rory Sayres, Ellery Wulczyn, Le~Hou, Kevin Clark, Stephen Pfohl, Heather Cole-Lewis, Darlene Neal, et~al.
\newblock Towards expert-level medical question answering with large language models.
\newblock \emph{arXiv preprint arXiv:2305.09617}, 2023.

\bibitem[SnorkelAI(2023)]{huggingfaceSnorkelaiSnorkelMistralPairRMDPODatasetDatasets}
SnorkelAI.
\newblock snorkelai/{S}norkel-{M}istral-{P}air{R}{M}-{D}{P}{O}-{D}ataset · {D}atasets at {H}ugging {F}ace --- huggingface.co.
\newblock \url{https://tinyurl.com/2ze38278}, 2023.
\newblock [Accessed 07-08-2024].

\bibitem[Thirunavukarasu et~al.(2023)Thirunavukarasu, Ting, Elangovan, Gutierrez, Tan, and Ting]{thirunavukarasu2023large}
Arun~James Thirunavukarasu, Darren Shu~Jeng Ting, Kabilan Elangovan, Laura Gutierrez, Ting~Fang Tan, and Daniel Shu~Wei Ting.
\newblock Large language models in medicine.
\newblock \emph{Nature medicine}, 29\penalty0 (8):\penalty0 1930--1940, 2023.

\bibitem[Toma et~al.(2023)Toma, Lawler, Ba, Krishnan, Rubin, and Wang]{toma2023clinical}
Augustin Toma, Patrick~R Lawler, Jimmy Ba, Rahul~G Krishnan, Barry~B Rubin, and Bo~Wang.
\newblock Clinical camel: An open expert-level medical language model with dialogue-based knowledge encoding.
\newblock \emph{arXiv preprint arXiv:2305.12031}, 2023.

\bibitem[Tran et~al.(2023)Tran, Glaze, and Hancock]{viethoangtranduong}
Hoang Tran, Chris Glaze, and Braden Hancock.
\newblock Iterative dpo alignment.
\newblock Technical report, Snorkel AI, 2023.

\bibitem[Tunstall et~al.()Tunstall, Beeching, Lambert, Rajani, Huang, Rasul, Bartolome, M.~Rush, and Wolf]{Tunstall_The_Alignment_Handbook}
Lewis Tunstall, Edward Beeching, Nathan Lambert, Nazneen Rajani, Shengyi Huang, Kashif Rasul, Alvaro Bartolome, Alexander M.~Rush, and Thomas Wolf.
\newblock {The Alignment Handbook}.
\newblock URL \url{https://github.com/huggingface/alignment-handbook}.

\bibitem[Tunstall et~al.(2023)Tunstall, Beeching, Lambert, Rajani, Rasul, Belkada, Huang, von Werra, Fourrier, Habib, Sarrazin, Sanseviero, Rush, and Wolf]{tunstall2023zephyr}
Lewis Tunstall, Edward Beeching, Nathan Lambert, Nazneen Rajani, Kashif Rasul, Younes Belkada, Shengyi Huang, Leandro von Werra, Clémentine Fourrier, Nathan Habib, Nathan Sarrazin, Omar Sanseviero, Alexander~M. Rush, and Thomas Wolf.
\newblock Zephyr: Direct distillation of lm alignment, 2023.

\bibitem[Vilares \& G{\'o}mez-Rodr{\'i}guez(2019)Vilares and G{\'o}mez-Rodr{\'i}guez]{vilares-gomez-rodriguez-2019-head}
David Vilares and Carlos G{\'o}mez-Rodr{\'i}guez.
\newblock {HEAD}-{QA}: A healthcare dataset for complex reasoning.
\newblock In \emph{Proceedings of the 57th Annual Meeting of the Association for Computational Linguistics}, pp.\  960--966, Florence, Italy, July 2019. Association for Computational Linguistics.
\newblock \doi{10.18653/v1/P19-1092}.
\newblock URL \url{https://www.aclweb.org/anthology/P19-1092}.

\bibitem[Wang et~al.(2020)Wang, Lo, Chandrasekhar, Reas, Yang, Burdick, Eide, Funk, Katsis, Kinney, Li, Liu, Merrill, Mooney, Murdick, Rishi, Sheehan, Shen, Stilson, Wade, Wang, Wang, Wilhelm, Xie, Raymond, Weld, Etzioni, and Kohlmeier]{wang-etal-2020-cord}
Lucy~Lu Wang, Kyle Lo, Yoganand Chandrasekhar, Russell Reas, Jiangjiang Yang, Doug Burdick, Darrin Eide, Kathryn Funk, Yannis Katsis, Rodney~Michael Kinney, Yunyao Li, Ziyang Liu, William Merrill, Paul Mooney, Dewey~A. Murdick, Devvret Rishi, Jerry Sheehan, Zhihong Shen, Brandon Stilson, Alex~D. Wade, Kuansan Wang, Nancy Xin~Ru Wang, Christopher Wilhelm, Boya Xie, Douglas~M. Raymond, Daniel~S. Weld, Oren Etzioni, and Sebastian Kohlmeier.
\newblock {CORD-19}: The {COVID-19} open research dataset.
\newblock In \emph{Proceedings of the 1st Workshop on {NLP} for {COVID-19} at {ACL} 2020}, Online, July 2020. Association for Computational Linguistics.
\newblock URL \url{https://www.aclweb.org/anthology/2020.nlpcovid19-acl.1}.

\bibitem[Wang et~al.(2022)Wang, Mishra, Alipoormolabashi, Kordi, Mirzaei, Arunkumar, Ashok, Dhanasekaran, Naik, Stap, Pathak, Karamanolakis, Lai, Purohit, Mondal, Anderson, Kuznia, Doshi, Patel, Pal, Moradshahi, Parmar, Purohit, Varshney, Kaza, Verma, Puri, Karia, Sampat, Doshi, Mishra, Reddy, Patro, Dixit, Shen, Baral, Choi, Smith, Hajishirzi, and Khashabi]{wang2022supernaturalinstructions}
Yizhong Wang, Swaroop Mishra, Pegah Alipoormolabashi, Yeganeh Kordi, Amirreza Mirzaei, Anjana Arunkumar, Arjun Ashok, Arut~Selvan Dhanasekaran, Atharva Naik, David Stap, Eshaan Pathak, Giannis Karamanolakis, Haizhi~Gary Lai, Ishan Purohit, Ishani Mondal, Jacob Anderson, Kirby Kuznia, Krima Doshi, Maitreya Patel, Kuntal~Kumar Pal, Mehrad Moradshahi, Mihir Parmar, Mirali Purohit, Neeraj Varshney, Phani~Rohitha Kaza, Pulkit Verma, Ravsehaj~Singh Puri, Rushang Karia, Shailaja~Keyur Sampat, Savan Doshi, Siddhartha Mishra, Sujan Reddy, Sumanta Patro, Tanay Dixit, Xudong Shen, Chitta Baral, Yejin Choi, Noah~A. Smith, Hannaneh Hajishirzi, and Daniel Khashabi.
\newblock Super-naturalinstructions: Generalization via declarative instructions on 1600+ nlp tasks, 2022.

\bibitem[Wang et~al.(2024)Wang, Ma, Zhang, Ni, Chandra, Guo, Ren, Arulraj, He, Jiang, Li, Ku, Wang, Zhuang, Fan, Yue, and Chen]{Wang2024ds}
Yubo Wang, Xueguang Ma, Ge~Zhang, Yuansheng Ni, Abhranil Chandra, Shiguang Guo, Weiming Ren, Aaran Arulraj, Xuan He, Ziyan Jiang, Tianle Li, Max Ku, Kai Wang, Alex Zhuang, Rongqi Fan, Xiang Yue, and Wenhu Chen.
\newblock {MMLU-Pro}: A more robust and challenging {Multi-Task} language understanding benchmark.
\newblock June 2024.

\bibitem[Wei et~al.(2022)Wei, Bosma, Zhao, Guu, Yu, Lester, Du, Dai, and Le]{wei2022finetuned}
Jason Wei, Maarten Bosma, Vincent~Y. Zhao, Kelvin Guu, Adams~Wei Yu, Brian Lester, Nan Du, Andrew~M. Dai, and Quoc~V. Le.
\newblock Finetuned language models are zero-shot learners, 2022.

\bibitem[Yim et~al.(2023)Yim, Fu, {Ben Abacha}, Snider, Lin, and Yetisgen]{aci-bench}
Wen{-}wai Yim, Yujuan Fu, Asma {Ben Abacha}, Neal Snider, Thomas Lin, and Meliha Yetisgen.
\newblock Aci-bench: a novel ambient clinical intelligence dataset for benchmarking automatic visit note generation.
\newblock \emph{Nature Scientific Data}, 2023.

\end{thebibliography}
\bibliographystyle{iclr2024_conference}

\newpage
\appendix

\section{Appendix}
\begin{table*}[!h]
%\centering
\begin{tabular}{lcc}
\toprule
% \textbf{Dataset}                   & \multicolumn{1}{c}{\textbf{\# Samples}} & \multicolumn{1}{c}{\textbf{Mixture ratio (\%)}} \\
% \midrule
% Medical domain &  &  \\ 
% ~~~MedMCQA \citep{pmlr-v174-pal22a} & 180,462 & 23.49 \\ 
% ~~~Medical Flashcards \citep{Han2023medalpaca} & 30,106 & 3.92 \\ 
% ~~~StackExchange \citep{h4stackexchange} & 64,246 & 8.36 \\ 
% ~~~MedQA (USMLE) \citep{jin2020disease} & 11,290 & 1.47 \\ 
% ~~~CORD-19 \citep{wang-etal-2020-cord} & 17,721 & 2.31 \\ 
% ~~~PubMedQA \citep{jin2019pubmedqa} & 499 & 0.06 \\ 
% ~~~HeadQA \citep{vilares-gomez-rodriguez-2019-head} & 2,657 & 0.35 \\ 
% ~~~MediQA \citep{Han2023medalpaca} & 1,950 & 0.25 \\ 
% ~~~SciQ \citep{SciQ} & 11,679 & 1.52 \\ 
% ~~~PubMed Causal \citep{Han2023medalpaca} & 2,169 & 0.28 \\ 
% ~~~OpenGPT & 66,026 & 8.59 \\ 
% ~~~MedQUAD \citep{BenAbacha-BMC-2019} & 14,553 & 1.89 \\ 
% ~~~MMLU \citep{hendryckstest2021} & 244 & 0.03 \\ 
% ~~~Niv2* \citep{wang2022supernaturalinstructions} & 11,447 & 1.49 \\ 
% ~~~Pubhealth \citep{kotonya2020explainable} & 9,804 & 1.28 \\ 
% ~~~Medical-Instruction \citep{medical-instruction-120k} & 120,000 & \\
% ~~~ACI-Bench \citep{aci-bench} & 87 & \\
% ~~~MTS-Dialog \citep{mts-dialog} & 2602 & \\
% Total & 424,853 & 55.29 \\ \midrule
% General domain &  & \\ 
% ~~~SlimOrca T0 \citep{OpenOrca, sanh2022multitask} & 109,235 & 14.22 \\ 
% ~~~SlimOrca Flan \citep{OpenOrca, longpre2023flan} & 109,169 & 14.21 \\ 
% ~~~SlimOrca CoT \citep{OpenOrca, wei2022finetuned} & 74,172 & 9.65 \\ 
% ~~~Ultrachat \citep{ding2023enhancing} & 50,953 & 6.63 \\ 
% Total & 343,529 & 44.71 \\ 
% \bottomrule
\textbf{Dataset}& \multicolumn{1}{c}{\textbf{\# Samples}} & \multicolumn{1}{c}{\textbf{Mixture ratio (\%)}} \\
\midrule
Medical domain & & \\
~~~MedMCQA \citep{pmlr-v174-pal22a} & 180,462 & 13.92 \\
~~~Medical Flashcards \citep{Han2023medalpaca} & 30,106 & 2.32 \\
~~~StackExchange\textsuperscript{\textdagger} \citep{h4stackexchange} & 64,246 & 4.96 \\
~~~MedQA (USMLE) \citep{jin2020disease} & 11,290 & 0.87 \\
~~~CORD-19 \citep{wang-etal-2020-cord} & 17,721 & 1.37 \\
~~~PubMedQA \citep{jin2019pubmedqa} & 499 & 0.04 \\
~~~HeadQA\textsuperscript{\textdaggerdbl} \citep{vilares-gomez-rodriguez-2019-head} & 2,657 & 0.20 \\
~~~MediQA \citep{Han2023medalpaca} & 1,950 & 0.15 \\
~~~SciQ \citep{SciQ} & 11,679 & 0.90 \\
~~~PubMed Causal \citep{Han2023medalpaca} & 2,169 & 0.17 \\
~~~OpenGPT \citep{cogstack-opengpt-sharegpt} & 66,026 & 5.09 \\
~~~MedQUAD \citep{BenAbacha-BMC-2019} & 14,553 & 1.12 \\
~~~MMLU\textsuperscript{\textdollar} \citep{hendryckstest2021} & 244 & 0.02 \\
~~~Niv2* \citep{wang2022supernaturalinstructions} & 11,447 & 0.88 \\
~~~Pubhealth \citep{kotonya2020explainable} & 9,804 & 0.76 \\
~~~Medical-Instruction \citep{medical-instruction-120k} & 120,000 & 9.26 \\
~~~ACI-Bench \citep{aci-bench} & 87 & 0.01 \\
~~~MTS-Dialog \citep{mts-dialog} & 2602 & 0.20 \\
Total & 952,942  & 73.50 \\ \midrule
General domain & & \\
~~~SlimOrca T0 \citep{OpenOrca, sanh2022multitask} & 109,235 & 8.43 \\
~~~SlimOrca Flan \citep{OpenOrca, longpre2023flan} & 109,169 & 8.42 \\
~~~SlimOrca CoT \citep{OpenOrca, wei2022finetuned} & 74,172 & 5.72 \\
~~~Ultrachat \citep{ding2023enhancing} & 50,953 & 3.93 \\
Total & 343,529 & 26.50 \\ 
\bottomrule
\multicolumn{3}{l}{\small \textsuperscript{\textdagger} The following categories were included: ``academia", ``bioinformatics'', ``biology", ``cogsci", ``fitness",} \\
\multicolumn{3}{l}{\small ``health".} \\
\multicolumn{3}{l}{\small \textsuperscript{\textdaggerdbl} Only samples in English were used. } \\
\multicolumn{3}{l}{\small \textsuperscript{\textdollar} The following subjects were included: ``anatomy", ``clinical knowledge", ``college medicine", ``nutrition",} \\
\multicolumn{3}{l}{\small ``medical genetics", ``professional medicine", ``college biology", ``high-school biology", ``virology", } \\
\multicolumn{3}{l}{\small  ``high-school psychology", ``human sexuality", ``human aging", and ``professional psychology".} \\
\multicolumn{3}{l}{\small * Samples from 47 tasks (from over 1,000 tasks) related to science, healthcare and medicine were included. }
\end{tabular}
\caption{Summary of subsets of the data used for fine-tuning the models. Note that medical-domain data correspond to 73.5\% of the entire dataset.}
\label{tab:traindata}
\end{table*}

\end{document}